# A secure home automation prototype built on raspberry-pi

Arya Tanmay Gupta, Humani Gupta, Muskan Sharma, Priyanka Khanna
Department of Computer Science, Ramanujan College (University of Delhi)

arya.tanmay.gupta@gmail.com

## Abstract

With the development of sensors, wireless mobile communication, embedded system, the technologies of the Internet of Things have been widely used in SmartMeter, public security, intelligent building and so on. Because of its huge market prospects, the Internet of Things has been paid close attention by several governments all over the world. IoT facilitates the seamless integration of wireless sensor networks.
In this paper, we present an IoT prototype that is built on Raspberry Pi and uses SMTP (simple mail transfer protocol) for communication. Through this device, we have proposed a communication system that is less complex and more secure. It integrates with any "thing" and makes it electronically communicable. We give an implementation of the prototyping system and system validation.

## I. INTRODUCTION

IoT is regarded as the third wave of Information Technology after the Internet and mobile communication network, which is characterized by more thorough sense and measure, more comprehensive interoperability and intelligence. Technologies of IoT can effectively facilitate the integration of material production and service management, the integration of physical and digital world.
Things are increasingly equipped with a large number of sensors, actuators, and communication devices (mobile devices, GPS devices and embedded computers). In particular, numerous things have possessed powerful sensing, networking, communication and data processing capabilities, and can communicate with other things or exchange info with external environments over various protocols.

The increasing demand and use of IoT technology have evolved various communication and security issues. The explosive growth of the requirement of the communication for information between machines raised concerns, such as the optimization of the human environment, the management of urban security, the improvement of living quality, and the effective management of production, [1] the "Internet of Things" (loT) is in great demand.
The advances in IoT have provided a promising opportunity which is Convenience. As more connected devices can handle more operations (lighting, switching, temperature etc.).
In 1999, MIT Auto-ID Labs first proposed the concept of the Internet of Things, which investigates to realize object localization and state recognition using wireless sensor networks(WSNs)and radio frequency identification technologies(RFIDs) [9], [24]. In 2005, International Telecommunication Union (ITU) released 'ITU Internet Report 2005: Internet of Things', formally proposed the concept of the Internet of Things, which noted that the ubiquitous Internet of Things communication era dawned, in which all objects in the world can exchange information via the networks actively [9], [25]. In 2009, IBM presented the "Smart-Planet" concept which aims to embed sensors in several physical objects such as power grid, railway, buildings, and make them smart by intelligent processing technologies [9], [26].
XaolinJia et al (2012) have discussed RFID Technology and its applications in IoT. They have described various applications and advantages of RFID and WSN. Various machines and prototypes like iMedBox [5], Opportunistic Large Array (OLA) [4], Complex Programmable Logic Device (CPLD) [6], and many others. They have used radio signals to establish communication between nodes or from nodes to a centralized system. [4] has studied SNR received by the nodes in Radio Communication. If we do communication directly in such a way, communication-related issues are potential.
Sachin Babar et al (2010) have presented the security requirements and security & threat taxonomy for IoT. They have insisted on going for a Trusted Platform Module which offers facilities for the secure generation of cryptographic keys, and limitations of their use, in addition to a hardware pseudorandom number generator.
We have objected to eliminating these issues, fulfilling the security requirements and improvising and simplifying the technological approaches being used to date.
We have built a prototype that works on Internet Protocol (IP) technology. It uses the IPV4 or IPV6 address, whichever the network supports, and communicates using the same network address. We have tried to keep the prototype's idea free from any special hardware requirements, other than it must get an IP address to start communication. This machine shows an idea of such a machine that is capable of being integrated with any "thing" which is trying to connect to the internet or any network to communicate and function through signals transferred through the same network. We call it Unitor. The communication that we present is secure and less sophisticated. The system overview of our prototype is given as figure 1.

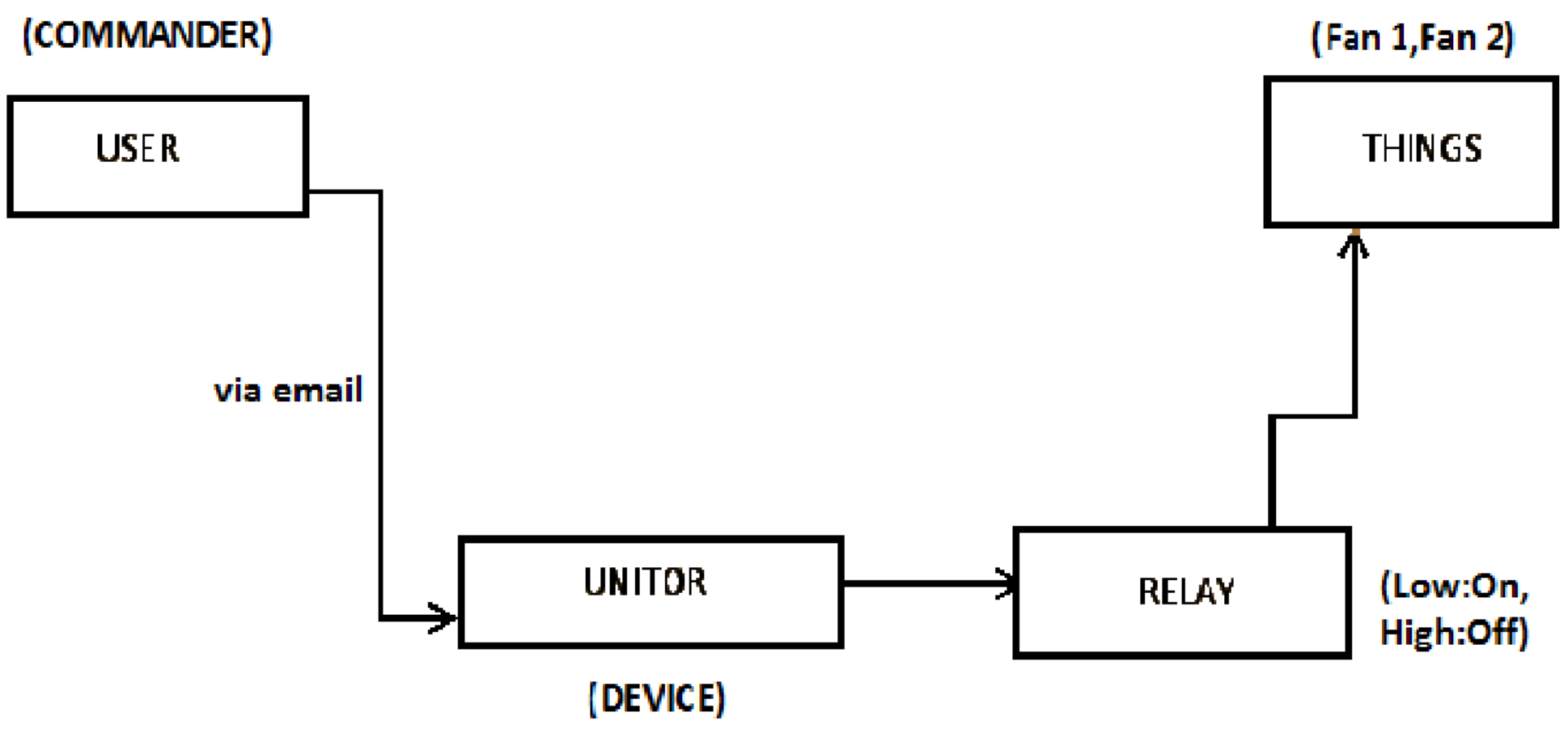

Figure 1. System overview

## II. PRIOR WORK AND COMPARISON

IoT Gateway was used with smart home by [9]. Smart home is a new type of house that is embedded with sensors, information home appliances, network communications and automation equipment, etc. in order to provide a comfortable, safe, green, convenient living environment. On one side, the in-home IoT Gateways play a very important role in interconnecting multiple smart devices together to form an in-home network and share resources and information among various home appliances. On the other side, in-home IOT Gateway also plays another role to connect the external networks to the in-home network and provide the access interface to the external networks.

Geng Yang et al have developed a health-IoT platform based on the integration of Intelligent Packaging, Unobtrusive Bio-Sensor, and Intelligent Medicine Box was built. It is an intelligent home-based health care system in which an intelligent medicine box (iMedBox) serves as a home healthcare gateway. It can deliver a variety of services such as on-site analysis, health social network, telemedicine, emergency and medication management services. In order to solve the medicine misuse problem, improve the pharmaceutical non-compliance situation, and make the daily task as easy and smart as possible, an intelligent medicine packaging (iMedPack) is built. IoT devices (e.g. wearable sensors, iMedPack, etc.) are connected to iMedBox.

[2] have discussed that RFID uses electromagnetic induction or electromagnetic propagation for the purpose of non-contact automatic identification of objects or humans. In wireless conditions, the affiliated EEPROM in the electronic label can be written and read by professional read-write equipment, RFID has the features against reproduction, combined encryption, and items, for the purpose of anti-counterfeiting.

[27] have developed Vehicular data cloud services in the IoT environment by integrating various devices such as sensors, actuators, controllers, GPS devices, mobile phones, and other internet access equipment and employing networking technologies, cloud computing, IoT, and middleware. This platform supports communication mechanisms and is able to

collect and exchange data among drivers, vehicles, and roadside infrastructures such as cameras and street lights.
M-health things (m-IoT) has been introduced recently and defined as the new concept that matches the functionalities of m-health and is mainly used for diabetes management. This device is based on IPV6 and 6LoWPAN protocols architectures [3], [15]. The non-invasive diabetes sensors from the patient are linked via IPV6 connectivity to a relevant healthcare provider. This device senses the glucose level of the patient and sends the data to the respective clinic. Thereby, physicians and nurses are able to check remotely patient's health status.
[28] have developed an electric bicycle system integrated with CAN-Bus protocol following a centralized architecture that communicates through Bluetooth BLE communication system.
[29] have presented a system that monitors and tracks various vital signs of a patient. Various medical sensors are integrated on an open and low-cost hardware platform. This system provides independence to the patient to send data related to his current medical situation without going to the medical centre. This system has been successfully tested on real patients.
[36] have proposed a oneM2M-based IoT system for Personal Healthcare Devices. They have shown that the conversion process (protocol conversion between ISO/IEEE 11073 protocol messages and oneM2M protocol messages is performed in this system) does not cause the system to suffer serious performance degradation, even when the number of Application Dedicated Node is quite large. They have also shown that the proposed algorithm can recover from faults on gateways in the oneM2M-based IoT system.

[30] have introduced Low-Power Wide Area Networks and discussed its advantages over established paradigms in terms of efficiency, effectiveness, and architectural design. They have insisted on the use of this technology for the typical Smart Cities applications. It is based on long-range radio links and star topology, the central node being the base station.
[31] have discussed the fog architecture and its usefulness in applications in IoT, 5G wireless systems and embedded AI.
[32] have discussed a worm that will spread rapidly from one lamp to its adjacent things using ZigBee wireless communication and their physical proximity. This threat will arise because of an increasing number of IoT devices currently. [32] have also opined that this chain reaction will fail if there are fewer than 15000 randomly located smart lamps in the city. They also suggested that this bug can be stopped by the ZigBee Light Link protocol, which is supposed to stop such attempts with a proximity test. Also, the use of cryptographic techniques can prevent such attacks.
An authentication system based on smartcards has been proposed in [33] where all the private information can be accessed securely by the registered users. Informal cryptanalysis confirms that this protocol is protected against all possible security threats. This protocol has been claimed to be more secure as compared to the protocols proposed by Xue et al [34] and Chuang et al [35].
The prototype which we will develop will be easy to handle with respect to networks. While WSN can realize the short distance communication among the objects by constructing wireless networks in ad-hoc manners. However, it is difficult to connect WSN and mobile communication networks or the internet with each other because it lacks uniform standardization in communication protocols and sensing technologies and the data from WNS cannot be transmitted in long-distance with the limitation of WNS's Transmission protocols. Therefore the goal is to handle heterogeneity between various sensor networks and mobile communication networks or the internet. Hence the prototype which we will develop will be able to handle the diversity of protocols in WNS and traditional telecommunication networks.
High-level security requirements in IoT are (1) resilience to attacks, (2) data authentication, (3) access control, (4) client privacy, (5) user identification, (6) secure storage, (7) identity management, (8) secure data communication, (9) availability, (10) secure network access, (11) secure content, (12) secure execution environment and (13) tamper resistance as suggested by Sachin Babar et al (2010).
Other communication requirements are (1) Ruggedized Communication Devices and Physical Media, (2) network topology, (3) resiliency and redundancy (4) network management (5) transportability (6) availability and reliability (7) time synchronization and accuracy (8) cybersecurity (9) maintenance. [23]
If we successfully fulfil these requirements, IOT can be implemented will any machine (or "thing") which can be communicated, controlled or monitored electronically. The prototype that we have developed fulfils these requirements.

# III. METHODOLOGY

## A. Basic Idea

The basic idea behind our prototype is to make any machine (or "thing") communicable, controllable or monitorable electronically.
We have tried to make the communication more secure and less sophisticated, both for the engineers and target users. Each thing will get an IP address which can be further used by it for node to node or node to user communication.

## B. The Architecture

Figure 2 shows the network architecture used in our prototype. This model is based on 4 layers: Physical & data link, network and application layers. We use SMTP at the application layer, IP at the network layer, and Ethernet at the physical and data-link layers.

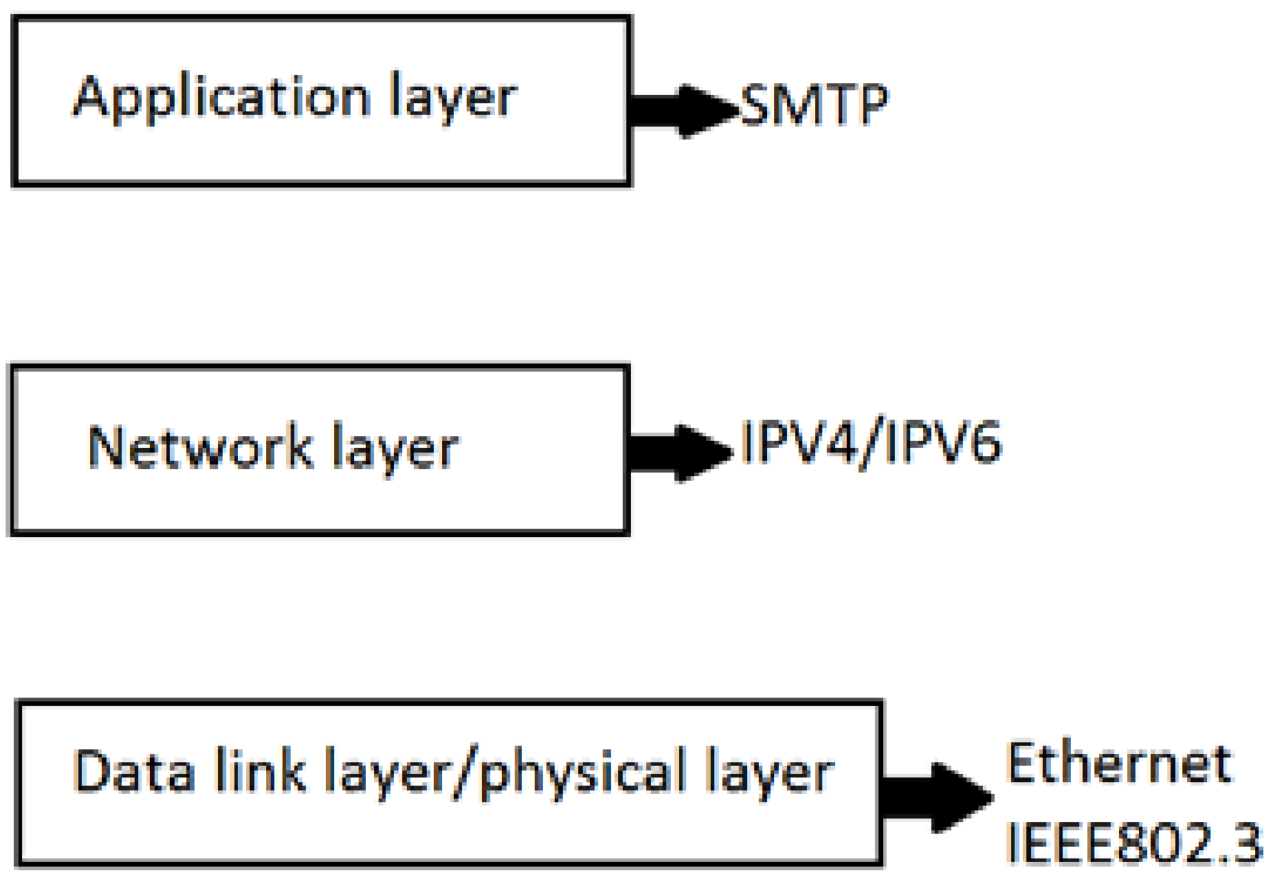

Figure 2. Network architecture.

To reduce the complexity of storage management, secure storage, and secure data communication we prefer not to establish communication directly. Instead of this, we have used the e-mail technology. We will, in this case, have a server to store data and all the security to the e-mail accounts. This model is shown in figure 3.

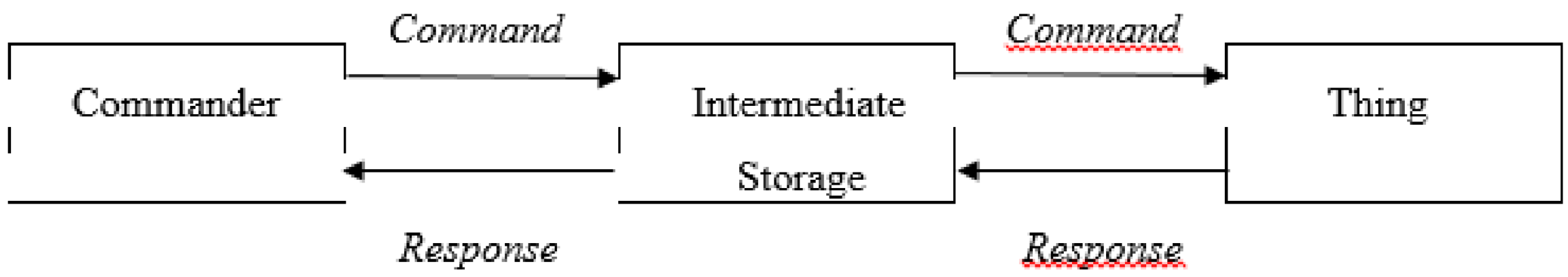

Figure 3. The communication model.

A better communication system will be made when we have various nodes working in coordination with each other, resulting in a smart system (for instance, a smart home or hospital), rather than having each node functioning independently. Internet of things is a collection of person-to machine and machine-to-machine communication systems [16], [18], [19].

Dong Chen et al (2011) have shown the random distribution of malicious and misbehaviour nodes in the simulations. This is shown in figure 6. If we establish communication between the nodes directly, the communication structure will be too complicated, as shown in figure 4.

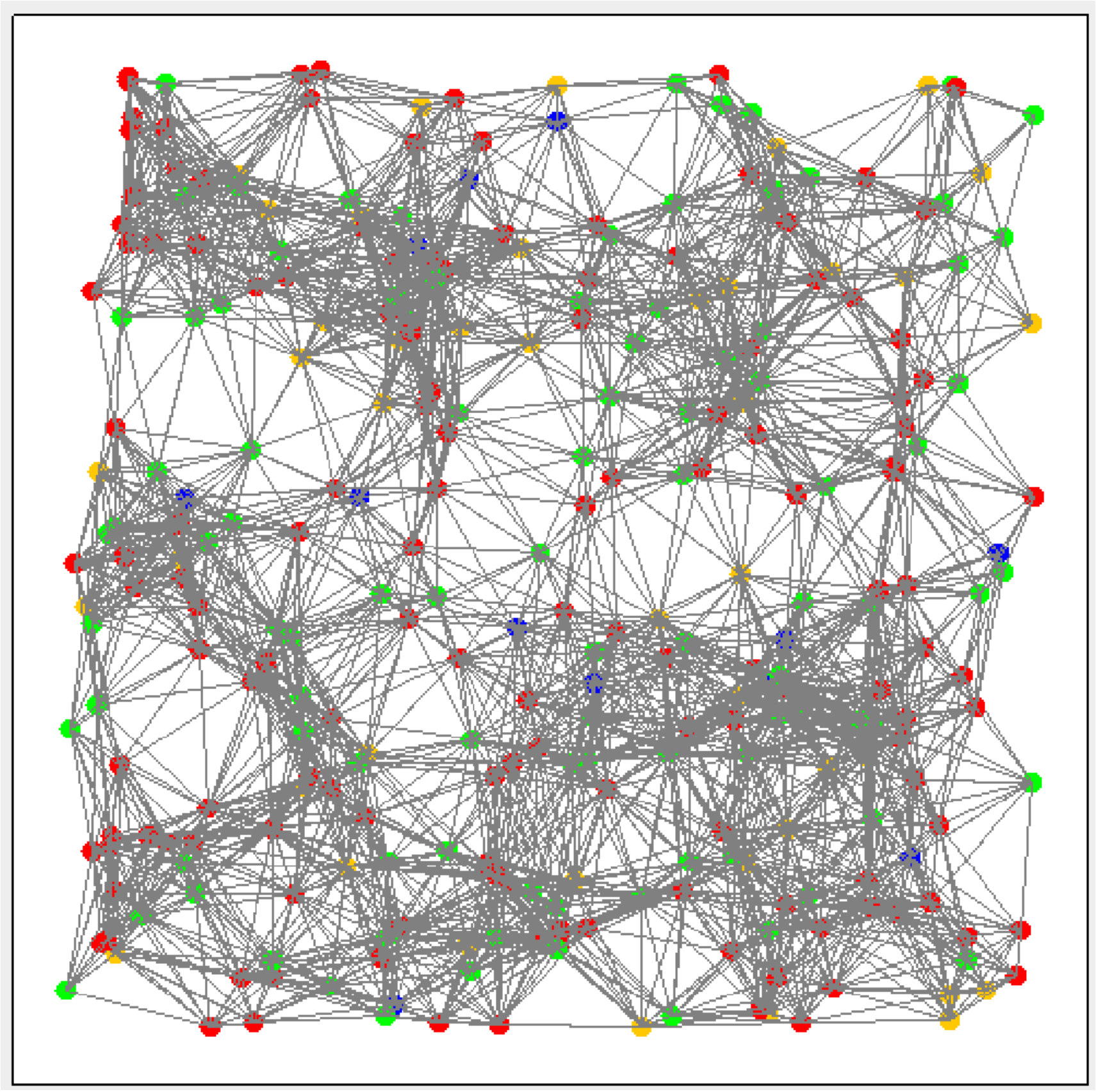

Figure 4. The random distribution of malicious and misbehaviour nodes in the simulations.

Email communication introduces an intermediate storage facility that will make the communication simpler, as the graph shown in figure 7. A similar architecture was implemented by [27] and [9] as in figure 5.

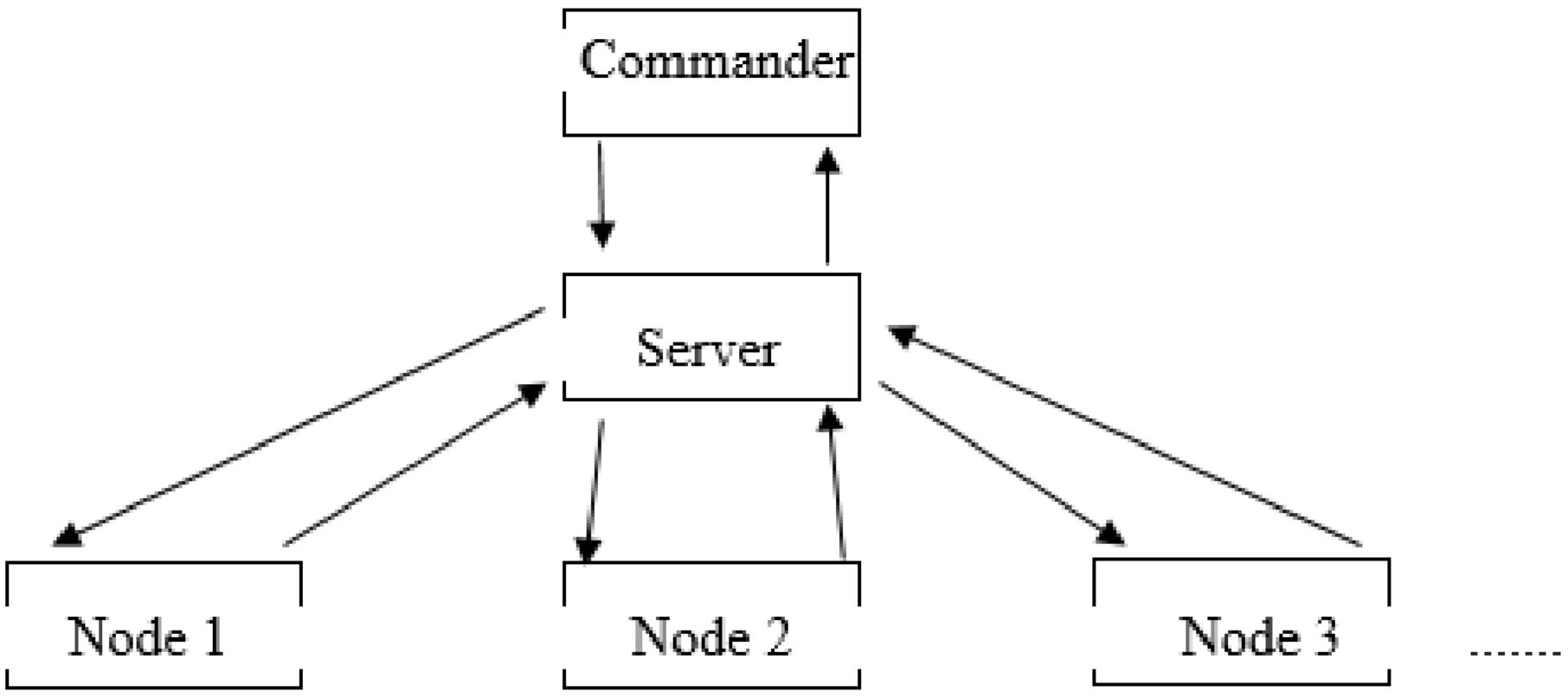

Figure 5. The structure of data transfer.

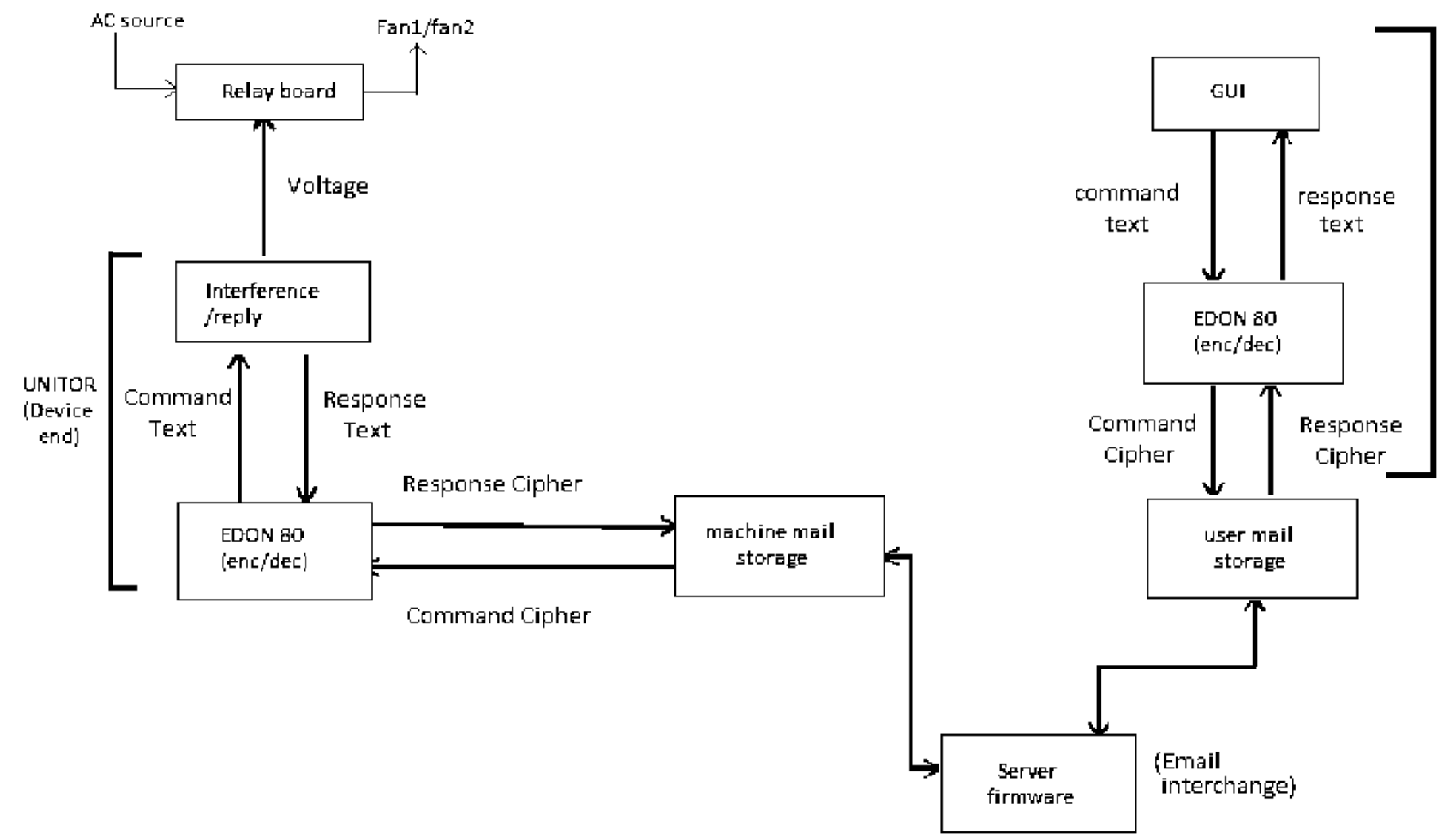

Figure 6. Communication architecture.

## C. Implementation and testing

For demonstration, we have built Unitor. Unitor is built using Raspberry Pi Model B+. It connects to the network through Ethernet (IEEE 802.1). The working program on Unitor is built in python. Raspberry Pi model 3 has 26 general-purpose input/output (GPIO) pins. So 26 "things" can be connected to Unitor.

We developed a GUI which takes commands from the user and sends to the "thing (s)". This software also contains the information of (1) the e-mail IDs of all the devices, (2) the "things" that the devices are integrated with, and (3) details about the list of commands related to the devices. It is built on Java Platform.

The user and the node(s) communicate in the method described in figure 6.

Both Unitor and the GUI use respective e-mail addresses to communicate to each other. Unitor ignores all the emails that are sent from some e-mail id, other than it is registered. It also keeps a check on the subject of e-mails sent to it.

The user clicks on one of the objects (or one of the "things" he wants to communicate with) and send commands or receive status(es) from Unitor about the object(s) it is connected with. Figure 7 shows the GUI and its possibilities of functions.

To overcome the issues of information to be hacked or cracked, we have used the keystream generated by Edon80 [21] algorithm to encrypt data. More details on the Edon80 encryption algorithm are given in section IV (A).

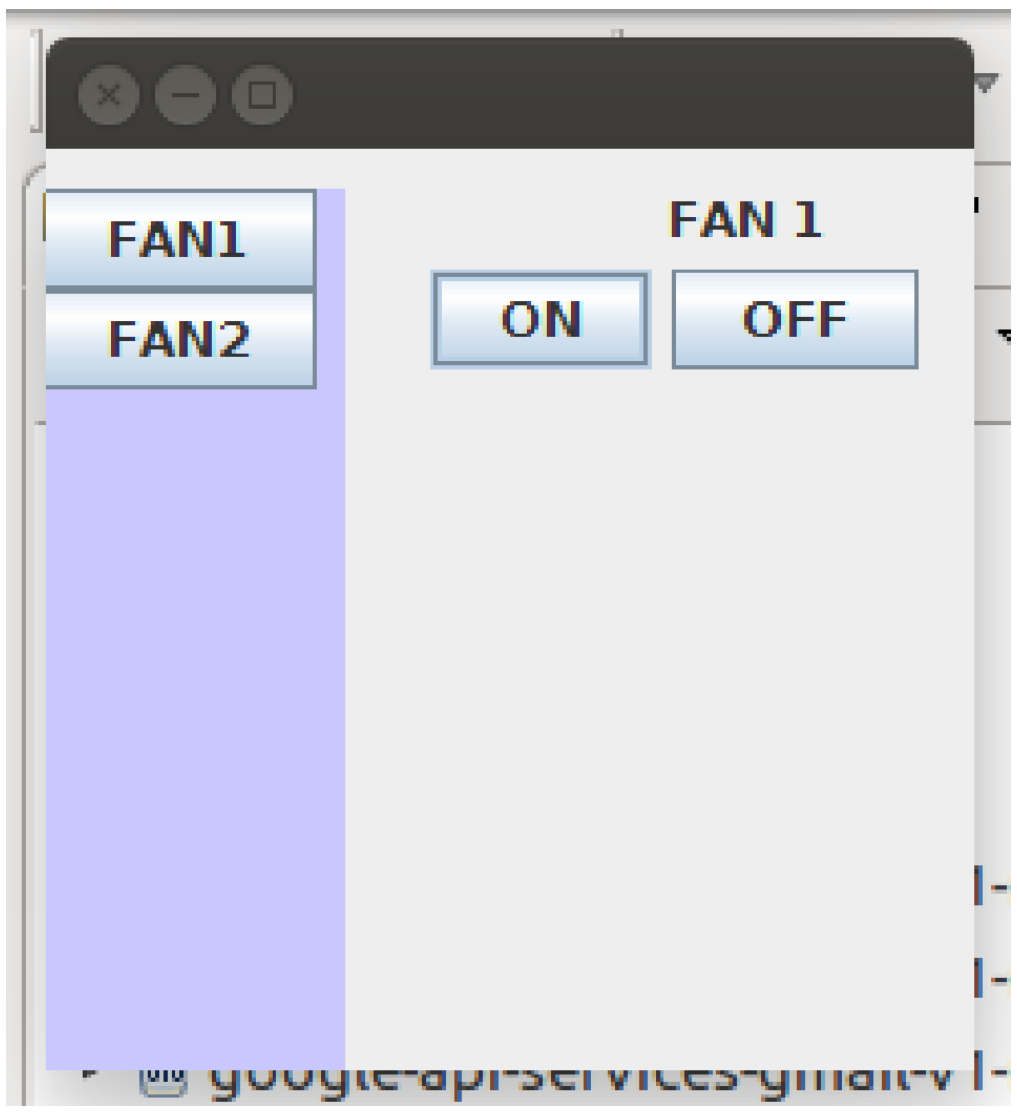

(a)

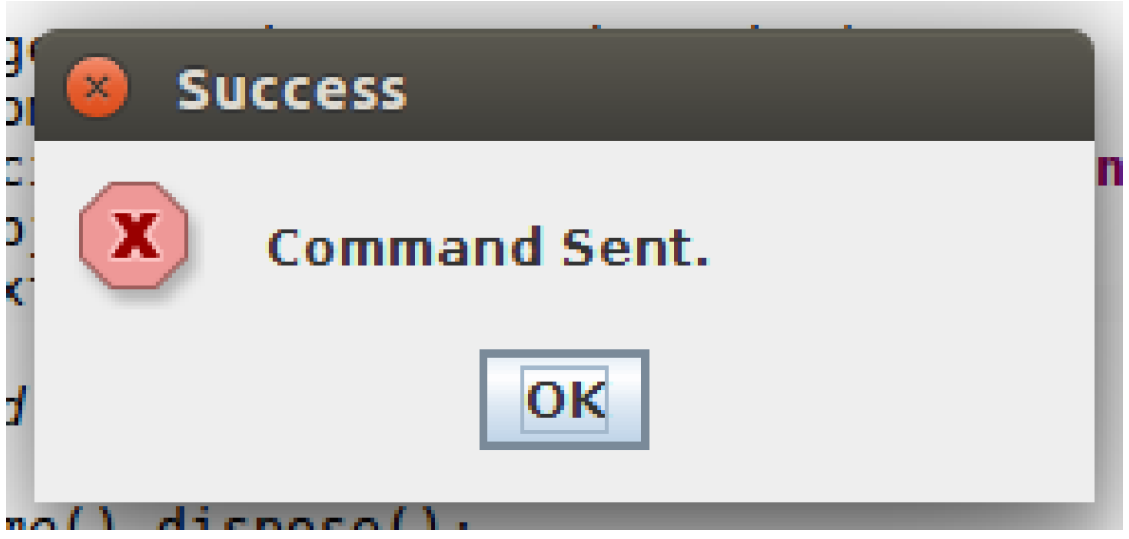

(b)

Figure 7. (a) The main window of the GUI. An example to command 2 fans. The left pane contains 2 options – "on" and "off" – available with "fan1". (b) On the success of the e-mail sent, the user is notified with "command sent" notice.

The implementation code is provided at [38].

## IV. BENEFITS

Some key benefits that Unitor renders are as follows:

- The time complexity of communication is precisely constant irrespective of the physical distance between the user and the device.
- It is a three-layer secure system as (a) It ignores all the e-mails sent by any other email-ids except the one(s) which are authenticated by Unitor, (b) if a hacker spoofs, it checks for the subject of the mail and ignores the unregistered one(s), and (c) we have encrypted the data that is being transferred during communication.
- We have presented a prototype that can connect with any number of devices as Raspberry Pi B+ has 26 GPIO pins which can be used to integrate with 26 devices.
- This design provides hardware and software platform independence using email

technology so the user can use a cheaper microcontroller other than Raspberry Pi.
By making minor changes in the design Unitor, phenomenal benefits can be rendered to the user.

## A. About the Encryption algorithm: Edon80

Edon80 is a binary additive stream cipher [20], [21]. It uses a key K which is of length 80 bits and an initial vector IV, which is of length 64 bits. This IV will be padded with 16 bits fixed value, that is 1 1 1 0 0 1 0 0 0 0 0 1 1 0 1 1. Using the key and IV generates a keystream which is of a variable length. That keystream generates cypher from the message. Figure 3 is a schematic representation of Edon80 as a binary additive stream cypher.

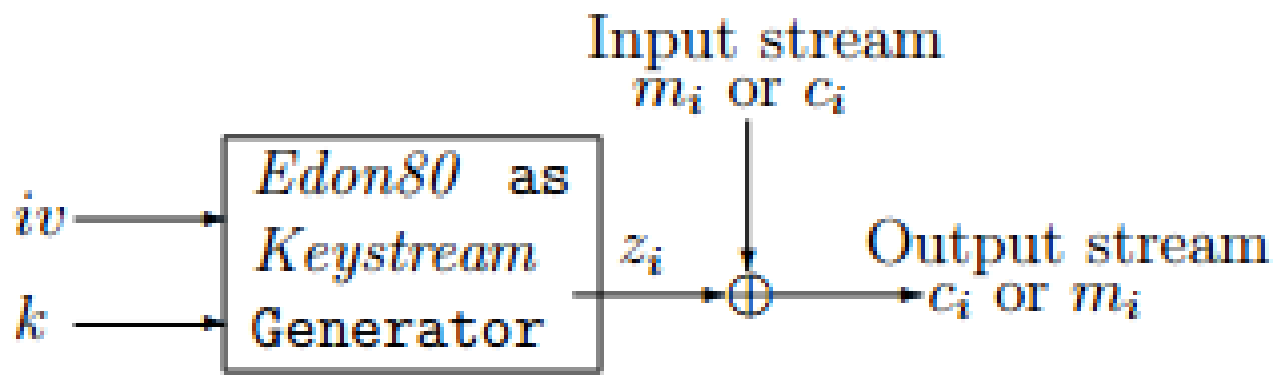

Figure 2: Schematic representation of Edon80 as a binary additive stream cypher.

We use 4 quasigroups of order 4 here, which are a total of 576 in number. Out of those, the developer has declared 64 quasigroups to be good to use in the algorithm. Out of those 64, quasigroups, the developer has used 4 distinct fixed quasigroups in the algorithm. Here we discuss the implementation process of the algorithm in MatLab.

### Edon 80 Algorithm

a. We have to input IV (length = 64 bits) and Key (length = 80 bits)
b. We have to modify IV to 80 bits by adding padding bits (1 1 1 0 0 1 0 0 0 0 0 1 1 0 1 1) to it.
c. Store the 4 quasigroups in an array. The four quasigroups as suggested by [21] are as follows.

| $\bullet_1$ | 1 | 2 | 3 | 4 |
|---|---|---|---|---|
| 1 | 1 | 3 | 2 | 4 |
| 2 | 3 | 2 | 4 | 1 |
| 3 | 2 | 4 | 1 | 3 |
| 4 | 4 | 1 | 3 | 2 |

| $\bullet_2$ | 1 | 2 | 3 | 4 |
|---|---|---|---|---|
| 1 | 2 | 4 | 1 | 3 |
| 2 | 1 | 2 | 3 | 4 |
| 3 | 3 | 1 | 4 | 2 |
| 4 | 4 | 3 | 2 | 1 |

| $\bullet_3$ | 1 | 2 | 3 | 4 |
|---|---|---|---|---|
| 1 | 3 | 2 | 1 | 4 |
| 2 | 2 | 3 | 4 | 1 |
| 3 | 4 | 1 | 3 | 2 |
| 4 | 1 | 4 | 2 | 3 |

| $\bullet_4$ | 1 | 2 | 3 | 4 |
|---|---|---|---|---|
| 1 | 4 | 3 | 2 | 1 |
| 2 | 2 | 1 | 4 | 3 |
| 3 | 1 | 4 | 3 | 2 |
| 4 | 3 | 2 | 1 | 4 |

d. Then we generate table 1 by mixing key and IV based on 80 times iteration of e-transformation based on key-based quasigroup operation (working quasigroup). The key is denoted by K = $k_1\, k_2\, k_3 \ldots$, and the IV is denoted by $v_1\, v_2\, v_3 \ldots$

e. The last row of table 1 is then used to generate table 2 according to the size of the input (message or cypher) using a pattern initial input string of quasigroup elements.

f. The keystream is obtained from table 2 and the input then xor with this keystream to get cypher/message.

**Algorithm:**

1. Start.
2. Input ***k*** = key (80 bits) and ***v*** = initial vector (length = 64 bits) in binary digit form.
3. Add padding values, i.e., 1 1 1 0 0 1 0 0 0 0 0 1 1 0 1 1, at the end of ***v***.
4. Convert the key ***k*** to quasigroup form $\boldsymbol{K} = K_1\, K_2\, K_3 \ldots K_{40}$.
5. Convert the initial vector ***v*** to quasigroup form $\boldsymbol{V} = V_1\, V_2\, V_3 \ldots V_{40}$.
6. Store $(Q, \bullet_1)$, $(Q, \bullet_2)$, $(Q, \bullet_3)$, $(Q, \bullet_4)$ in array.
7. Calculate working quasigroup operation $*_i = \begin{cases} \bullet_{K_i} & 1 \le i \le 40 \\ \bullet_{K_{i-40}} & 41 \le i \le 80 \end{cases}$
8. Generate table 1 (table 2 of Edon80 keystream mode) using the following algorithm:
   i. $t_{1,1} = v_{40} *_1 K_1$

| | | |
|---|---|---|
| ii. | $t_{1,j} = t_{1,j-1} *_1 K_j$ | $2 \leq j \leq 40$ |
| iii. | $t_{1,j} = t_{1,j-1} *_1 V_{j-40}$ | $41 \leq j \leq 80$ |
| iv. | $t_{i,1} = V_{40-i} *_i t_{i-1,1}$ | $2 \leq i \leq 40$ |
| v. | $t_{i,1} = K_{80-i} *_i t_{i-1,1}$ | $40 \leq i \leq 80$ |
| vi. | $t_{i,j} = t_{i,j-1} *_i t_{i-1,j}$ | $2 \leq i \leq 80, 2 \leq j \leq 80$ |

9. ***a*** = the last row of table1.
10. Input bit size of keystream.
11. Generate table 2 (table 3 of Edon80 keystream mode) using the following algorithm (; the row length is the bit size of keystream):

| | | |
|---|---|---|
| i. | $a_{1,1} = a_1 *_1 1$ | |
| ii. | $a_{0,j} = a_{0,j-1} *_1 ((j-1) \bmod 4 + 1)$ | $2 \leq j$ |
| iii. | $a_{i,1} = a_i *_i a_{i-1,1}$ | $2 \leq i \leq 80$ |
| iv. | $a_{i,j} = a_{i,j-1} *_i a_{i-1,j}$ | $2 \leq i \leq 80, 2 \leq j$ |

12. ***x*** = last row of table 2.
13. ***finalKey*** = an array of each element in x at even positions.
14. ***f*** = convert ***finalKey***, the keastream in bits form.
15. Choose the type of input message – alphabetical, hexadecimal, quasigroup form, or binary form.
16. Input message.
17. ***m*** = Convert the message into bits form correspondingly.
18. Xor each element of ***m*** with each element of ***f*** at corresponding positions and store in cypher.
19. The array cypher is the ciphertext.
20. Convert the ciphertext in required form – hexadecimal, quasigroup form, or unchanged to binary form.
21. Stop.

In this algorithm, step 18 has been used to generate cypher from original plaintext using the keystream by xor operation. Using the same method, the cypher can be converted back to the original plaintext. This is possible due to the reversibility of the xor operation. Edon80 is a robust encryption algorithm the generation of keystream is a time-taking process, but once a keystream is generated, we can use it again and again to encrypt and decrypt the data, which has linear complexity.

In our prototype, the keystream that we used uses the following matrices to generate the keystream from the Edon80 algorithm:

V = 0000000000000000000000000000000000000000000000000000000000000000

K =

100000000000000000000000000000000000000000000000000000000000000000000000000

00000

Quasigroup 1: 1243243143123124

Quasigroup 2: 1243243143123124

Quasigroup 3: 1243243143123124

Quasigroup 4: 1243243143123124

Edon80 has passed the NIST (National Institute of Standards and Technology [37]) test [22]. The research related to Edon80 and its test as per the NIST protocol has been presented in detail in [20] along with all the statistics. The keystream generated by this algorithm has been used to encrypt the text being sent in the communication from both sides.

This will make the data that we use in communication more secure. Unitor ignores any e-mail which is decrypted to form an unknown string, not resembling any string from the protocol set.

## B. More Improvement

Quasigroups can be used according to the date and altered in real-time, for example, every day at 12:00 AM, we can change the quasigroups in the application. If the date is 09-Apr-1996

- Use quasigroup number 9 from the lexicographical list of the quasigroups of order 4 (the total number of distinct quasigroups of order 4 being 576) as the first quasigroup in the algorithm (referring to the date);
- Use quasigroup number 4 as the second quasigroup (referring the month);
- Use quasigroup number 19 as the third quasigroup (referring the year – first two digits); and
- Use quasigroup number 96 as the fourth quasigroup (referring the year – third and fourth digits).

Using Wi-Fi instead of Ethernet, we can get the independence of motion. Using GSM technology instead of Ethernet, we can get the geo-location of the device using a separate GPS module.

If there is some error in connectivity we will not be able to communicate. In this situation, we can add a switch that will decide whether to use the thing by direct power or through Unitor, shown in figure 8.

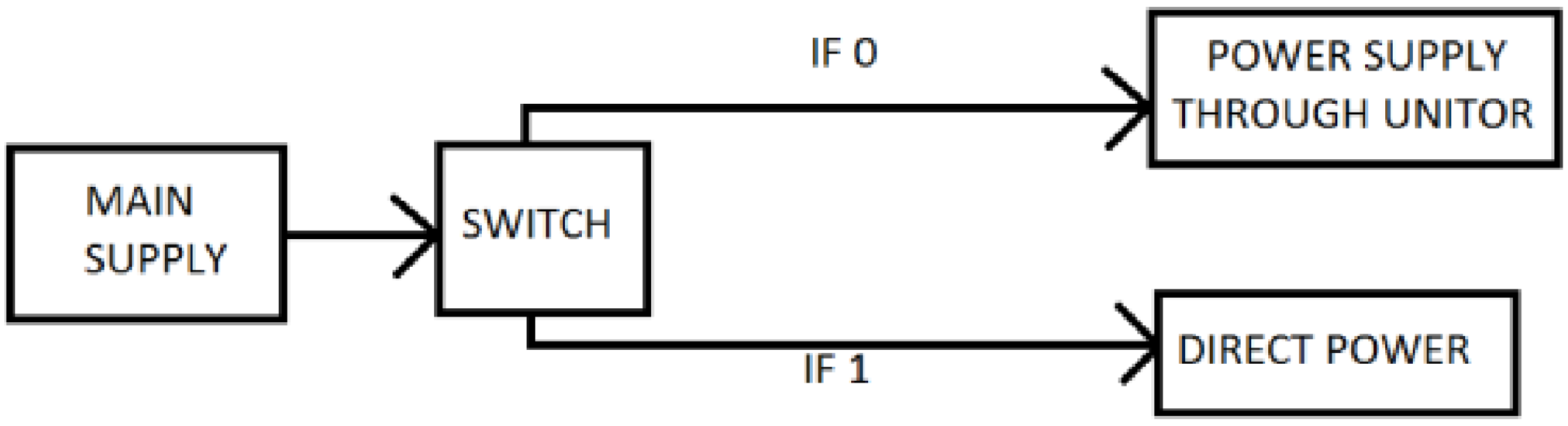

Figure 8. A design of electrical connections to handle connectivity faults.

Proposing this model, we insist on making the machine obtain an IP address first and not do the communication directly through ZigBee or RFIDs. We insist on making it compulsory to use the Internet Protocol at the network layer. If we target for the IP, we have a wide range of protocols above and below the network layer, for example, Wi-Fi at the physical layer, or GPRS of GSM and the SMB protocol at the application layer. We can even use the IP address directly for communication.

Hence, we increase the dimensions of independence of protocols and devices and enable the machine to access the functionalities that can be obtained only through the Internet Protocol, and make our "thing" actually use the Internet, as the name IOT suggests, for communication.

This is expected to make the lives of people better by providing more reliable, secure and less complex communication. It can be used with any devices or appliances from those at a scientific laboratory to a home for controlling and monitoring purposes. It can not only just turn a machine on or off, but by the functionalities that microcontrollers provide, we can read or generate pulses too. So it can be used with any electronic device.

Using this prototype, various devices can communicate with each other, providing more intelligent systems free from any human intervention as suggested by [16], [18], [19].

## CONCLUSION

This paper describes a prototypic implementation of an embedded system that connects any "thing" to the network through Ethernet. It communicates via e-mail. We have used raspberry pi to build Unitor. We have developed a GUI which takes command from a user and sends it to the Unitor.

We have combined all the best available approaches for security, communication, cost-effectiveness and develop a unified platform for node to user and node to node communication for the best possible automated systems by giving independence of hardware and software to the users. We have successfully implemented and tested this system.

## REFERENCES

[1] M. Yun, W. Hubei, "Research on the Architecture and Key Technology of Internet of Things (loT) Applied on Smart Grid", International Conference on Advances in Energy Engineering, 978-1-4244-7830-9/10 IEEE, 2010.

[2] X. Jia, Q. Feng, T. Fan and Q. Lei, "RFID Technology and its applications in Internet of Things (IoT)", 978-1-4577-1415-3/12 IEEE, 2010.

[3] R. S. H. Istepanian, A. Sungoor, A. Faisal and N. Philip, "INTERNET OF M-HEALTH THINGS "m-IOT"", IET Seminar on Assisted Living, IEEE, 2011.

[4] V. M. Rohokale, N. R. Prasad and R. Prasad, "A Cooperative Internet of Things (IoT) for Rural Healthcare Monitoring and Control", 978-1-4577-0787-2/11 IEEE, 2011.

[5] G. Yang, M. Mäntysalo, X. Zhou, Z. Pang, L. D. Xu, S. Kao-Water, Q. Chen and L.

Zheng, “A Health-IoT Platform Based on the Integration of Intelligent Packaging, Unobtrusive Bio-Sensor and Intelligent Medicine Box”, DOI 10.1109/TII.2014.2307795, IEEE Transactions on Industrial Informatics, 2013.
[6] Q. Chi, H. Yan, C. Zhang, Z. Pang and L. D. Xu, “A Reconfigurable Smart Sensor Interface for Industrial WSN in IoT Environment”, IEEE TRANSACTIONS ON INDUSTRIAL INFORMATICS, VOL. 10, NO. 2, MAY 2014.
[7] T. Yashiro, S. Kobayashi, N. Koshizuka and K. Sakamura, “An Internet of Things (IoT) Architecture for Embedded Appliances”, IEEE R10- HTC2013, Aug. 2013.
[8] F. Xia, L. T. Yang, L. Wang and A. Vinel, “Internet of Things”, Int. J. Commun. Syst.; 25:1101–1102, published online in Wiley Online Library (wileyonlinelibrary.com). DOI: 10.1002/dac.2417, 2012.
[9] Q. Zhu, R. Wang, Q. Chen, Y. Liu and W. Qin, “IOT Gateway: BridgingWireless Sensor Networks into Internet of Things”, 978-0- 7695-4322-2/10, DOI 10.1109/EUC.2010.58, IEEE, 2010.
[10] F. Tao, Y. Zuo, L. D. Xu and L. Zhang, “IoT-Based Intelligent Perception and Access of Manufacturing Resource Toward Cloud Manufacturing”, IEEE TRANSACTIONS ON INDUSTRIAL INFORMATICS, VOL. 10, NO. 2, MAY 2014.
[11] Y. J. Fan, Y. H. Yin, L. D. Xu, Y. Zeng and F. Wu, “IoT-Based Smart Rehabilitation System”, IEEE TRANSACTIONS ON INDUSTRIAL INFORMATICS, VOL. 10, NO. 2, MAY 2014.
[12] S. Babar, P. Mahalle, A. Stango, N. Prasad and R. Prasaad, “Proposed Security Model and Threat Taxonomy for the Internet of Things (IoT)”, N. Meghanathan et al. (Eds.): CNSA 2010, CCIS 89, pp. 420–429, 2010, Springer-Verlag Berlin Heidelberg 2010.
[13] M. Yun and B. Yuxin, “Research on the Architecture and Key Technology of Internet of Things (loT) Applied on Smart Grid”, 978-1- 4244-7830-9/10, IEEE, 2010.
[14] A. Iera, C. Floerkemeier, J. Mitsugi and G. Morabito, “THE INTERNET OF THINGS”, IEEE Wireless Communications (editorial), 2010.
[15] R. S. H. Istepian, H. Hu, N. Y. Philip and A. Sungoor, “The Potential of Internet of m-health Things “m-IoT” for Non-Invasive Glucose level Sensing”, 978-1-4244-4122-8/11, IEEE, 2011.
[16] S. D. T. Kelly, N. K. Suryadevara and S. C. Mukhopadhyay, “Towards the Implementation of IoT for Environmental Condition Monitoring in Homes”, IEEE SENSORS JOURNAL, VOL. 13, NO. 10, OCT. 2013.
[17] D. Chen, G. Chang, D. Sun, J. Li, J. Jia and X. Wang, “TRM-IoT: A Trust Management Model Based on Fuzzy Reputation for Internet of Things”, ComSIS Vol. 8, No. 4, DOI: 10.2298/CSIS110303056C, Oct. 2011.
[18] H. Sundmaeker, P. Guillemin, P. Friess, S. Woelffle, “Vision and Challenges for Realizing the Internet of Things”. Luxembourg, Germany: European Union, 2010, ISBN 9789279150883.
[19] “Internet 3.0: The Internet of Things”, Analysys Mason, Singapore, 2010. Available on http://www.theinternetofthings.eu/content/internet-30-internet-things-new-report-analysys-mason. Referred on Nov. 8, 2016.

[20] A. T. Gupta, “Development of Cryptographic Schemes using Quasigroups”, SAG, DRDO and Ramanujan College (University of Delhi), 2015. Available at https://github.com/ATGupta/Development-of-Quasigroup-based-Cryptographic-Schemes/blob/master/Report.pdf.
[21] D. Gligoroski, S. Markovski and S. J. Knapskog, “The Stream Cipher Edon80”. Conference paper DOI: 10.1007/978-3-540-68351-3_12, New Stream Cipher Designs, LNCS 4986, pp. 152–169, Springer-Verlag Berlin Heidelberg, 2008.
[22] “NIST.gov - Computer Security Division - Computer Security Resource Center”. Available on http://csrc.nist.gov/groups/ST/toolkit/rng/documentation_software.html.
[23] “Communication Network requirements”. Available on http://belden.picturepark.com/Website/Download.aspx?DownloadToken=63b8a52b-be50-4429-9a67-c283470e40f7&Purpose=AssetManager&mime-type=application/pdf.
[24] I. Bose and R. Pal, ”Auto-ID: managing anything, anywhere, anytime in the supply chain”, Communications of the ACM, Vol. 48, No. 8, pp. 100-106, Aug. 2005.
[25] ITU, “ITU Internet Reports 2005: The Internet of Things”, The Internet of Things, Nov. 2005.
[26] Q. Liu, L. Cui, HM. Chen”, Key technologies and applications of Internet of Things”, Computer Science, Vol. 37, No. 6, 2010.
[27] W. He, G. Yan and L. D. Xa, “Developing Vehicular Data Cloud Services in the IoT Environment”.
[28] J. R. Herrero, G. Villarrubia, A. L. Barriuso, D. Hernández, Á. Lozano, M. A. D. L. S. González (2015). Wireless controller and smartphone based interaction system for electric bicycles. ADCAIJ: Advances in Distributed Computing and Artificial Intelligence Journal, Salamanca, Vol. 4, No. 4.
[29] D. H. D. L. Iglesia, G. V. González, A. L. Barriuso, Á. L. Murciego, J. R. Herrero (2015). Monitoring and analysis of vital signs of a patient through a multi-agent application system. ADCAIJ: Advances in Distributed Computing and Artificial Intelligence Journal, Salamanca, Vol. 4, No. 3.
[30] M. Centenaro, L. Vangelista, A. Zanella, M. Zorzi (2016). Long-range communications in unlicensed bands: the rising stars in the IoT and smart city scenarios. IEEE Wireless Communications, Volume: 23, Issue 5.
[31] M. Chiang, T. Zhang (2016). Fog and IoT: An Overview of Research Opportunities. IEEE Internet of Things Journal, Volume 3, Issue 6.
[32] E. Ronen, A. Shamir, A. O. Weingarten, C. O’Flynn (2017). IoT Goes Nuclear: Creating a ZigBee Chain Reaction. IEEE Symposium on Security and Privacy (SP).
[33] R. Amin, N. Kumar, G. P. Biswas, R. Iqbal, V. Chang (2018). A light weight authentication protocol for IoT-enabled devices in distributed Cloud Computing environment. Future Generation Computer Systems, Volume 78, Part 3.
[34] K. Xue, P. Hong, C. Ma (2014). A lightweight dynamic pseudonym identity based authentication and key agreement protocol without verification tables for multi-server architecture” Journal of Computer and System Sciences 80 195-206.
[35] M. -C. Chuang, and M. C. Chen (2014). An anonymous multi-server authenticated key

agreement scheme based on trust computing using smartcards and biometrics. Expert Systems with Applications, vol. 41, no. 4(part 1).
[36] M. W. Woo, J. W. Lee, K. H. Park (2018). A reliable IoT system for Personal Healthcare Devices. Future Generation Computer Systems, Volume 78, Part 2.
[37] https://www.nist.gov/
[38] https://github.com/ATGupta/Unitor.